\documentclass[runningheads]{llncs}
\usepackage{graphicx}
\usepackage{booktabs} 
\usepackage{algorithm,algpseudocode}
\usepackage{placeins}
\usepackage{textcomp}
\usepackage[misc,geometry]{ifsym} 
\usepackage{url}

\urldef{\mailsa}\path|{alexander.hagg, alexander.asteroth}@h-brs.de|    
\urldef{\mailsb}\path|{m.preuss,t.h.w.baeck}@liacs.leidenuniv.nl|   

\begin{document}
\title{An Analysis of Phenotypic Diversity in Multi-Solution Optimization\thanks{This work received funding from the German Federal Ministry of Education and Research (BMBF) (grant agreement no. 03FH012PX5)}}
%
%
\author{Alexander Hagg\inst{1,2}\textsuperscript{(\Letter)}\orcidID{0000-0002-8668-1796} \and
	Mike Preuss\inst{2}\orcidID{0000-0003-4681-1346} \and
	Alexander Asteroth\inst{1}\orcidID{0000-0003-1133-9424} \and
	Thomas B\"ack\inst{2}\orcidID{0000-0001-6768-1478}}

\authorrunning{A. Hagg et al.}
%
\institute{Bonn-Rhein-Sieg University of Applied Sciences, Sankt Augustin, Germany\\\mailsa \and
	Leiden Institute of Advanced Computer Science, Leiden University, Leiden, The Netherlands\\\mailsb}
\maketitle              
\begin{abstract}
More and more, optimization methods are used to find diverse solution sets. We compare solution diversity in multi-objective optimization, multimodal optimization, and quality diversity in a simple domain. We show that multiobjective optimization does not always produce much diversity, multimodal optimization produces higher fitness solutions, and quality diversity is not sensitive to genetic neutrality and creates the most diverse set of solutions. An autoencoder is used to discover phenotypic features automatically, producing an even more diverse solution set with quality diversity. Finally, we make recommendations about when to use which approach.

\end{abstract}
%
%
%

\section{Introduction}
\label{sec:1}

With the advent of 3D printing and generative design, a new goal in optimization is emerging. Having the option of choosing from different solutions that are good enough to fulfill a task can be more effective than being guided by single-solution algorithms. The optimization field should aim to understand how to solve a problem in different ways.

Three major paradigms for multi-solution optimization exist. The major difference between multi-objective optimization (MOO), multimodal optimization (MMO) and quality diversity (QD) is the context in which solution diversity is maintained. In MOO the goal is to find the Pareto set, which represents the trade-offs between multiple criteria. MMO finds solutions that cover the search space as well as possible. QD finds combinations of phenotypic features to maximize the variation in solutions' expressed shape or behavior - a new focus in evolutionary optimization~\cite{Pugh2016}.

\noindent We analyze the diversity of solution sets in the three paradigms and introduce a new niching method that allows comparing genetic and phenotypic diversity (Section~\ref{sec:2}). State of the art diversity metrics (Section~\ref{sec:3}) are used in a new problem domain (Section~\ref{sec:4}) to evaluate all paradigms (Section~\ref{sec:5}) after which we make recommendations when to use which approach (Section~\ref{sec:6}). 

\section{Diversity in Optimization}
\label{sec:2}

The intuitive understanding of diversity assumes that there are more ways to ``do'' or to ``be'' something and involves the concepts of \textit{dissimilarity} and \textit{distance}. Evidence can be found in the large number of approaches and metrics, and the lack of agreement in when to use which one. This section gives an overview over three paradigms that have arisen in the last decades.

Finding solutions that are diverse with respect to objective space has been a paradigm since the 1970s. Multi-objective optimization tries to discover the Pareto set of trade-off solutions with respect to two or more objectives. The method has no control over the diversity of genomes or their expression other than the expectation that trade-offs require different solutions. The most successful method is the Non-dominated Sorting Genetic Algorithm (NSGA-II)~\cite{Deb2002}.


The first ideas to use genetic diversity in optimization were not used to find different solutions, but to deal with premature convergence to local optima. The concept of \textit{niching} was integrated into evolutionary optimization by introducing sharing and crowding~\cite{holland1975,Jong1975}. In the 1990s, multi-local or multimodal optimization came into focus. This paradigm has the explicit goal to find a diverse set of high quality locations in the search space, based on a single criterion. Various algorithms have been introduced, like basin hopping~\cite{Wales}, topographical selection~\cite{torn1992topographical}, nearest-better clustering~\cite{preuss2012improved} and restarted local search (RLS)~\cite{Posik2012}. 

The introduction of novelty search~\cite{Lehman2011a} led to studying the search for novel, non-optimal solutions. QD, reintroducing objectives~\cite{Cully2015,Lehman2011}, finds a diverse set of high quality optimizers by performing niching in phenotypic space. In applications for developing artificial creatures and robot controller morphologies~\cite{Cully2015,Lehman2011}, QD only allows solutions that belong the same phenotypic niche to compete. To this end it keeps track of an archive of niches. Solutions are added to the archive if their phenotype is novel enough or better than that of a similar solution. 

This work does not aim at giving an exhaustive overview over all methods, for which we refer to some of the many survey papers~\cite{Cully2017,Fernandes2013,Posik2012,Tian2017,Tian2019,Wang2017b}. We consciously choose not to talk about methods that combine ideas from the three paradigms, but rather compare the three paradigms in their ``purest'' form.

\subsection{Niching with Voronoi Tessellation}

To remove variations in the search dynamics when comparing different algorithms, we introduce a niching variant using ideas from Novelty Search with Local Competition (NSLC)~\cite{Lehman2011} and CVT-Elites~\cite{Vassiliades}. Voronoi-Elites (VE) accepts all new solutions until the maximum number of archive bins is surpassed (Alg.~\ref{alg1}). Then the pair of elites that are phenotypically closest to each other are compared, rejecting the worst-performing.  An example archive is shown in Fig.~\ref{img:autove} at step five). By locating selection pressure on the closest solutions, VE tries to equalize the distances between individuals. The generators of the Voronoi cells do not have to coincide with the centroids, like in CVT-Elites, and the boundaries of the archive are not fixed. VE can be used to compare archive spaces of different dimensionality. When the genetic parameters are used as archive dimensions, VE behaves like an MMO algorithm by performing niching in genetic space. When we use phenotypic descriptors, VE behaves like a QD algorithm.

\begin{algorithm}
	\caption{Voronoi-Elites}
	\label{alg1}
	\begin{algorithmic}
		\State \textbf{Initialize} population
		\For {iter 1 to n}
		\State \textbf{Select} parents $\mathcal{P}$ randomly
		\State \textbf{Mutate} $\mathcal{P}$ using normal distribution to create offspring $\mathcal{O}$
		\State \textbf{Evaluate} performance and descriptors of $\mathcal{O}$
		\State \textbf{Add} $\mathcal{O}$ to archive $\mathcal{A}$
		\While {$|\mathcal{A}| > maxSize$}
		\State \textbf{Find} pair in $\mathcal{A}$ with smallest distance
		\State \textbf{Remove} individual (in pair) with lowest fitness
		\EndWhile
		\EndFor
	\end{algorithmic}
\end{algorithm}

\subsection{Related Work}

A number of survey and analysis articles have appeared in the last decade. In~\cite{Fernandes2013} a taxonomy for diversity in optimization was introduced. \cite{Wessing2016} investigates how genetically diverse solution sets in MOO are found and shows that quality indicators used in MOO can be applied to MMO. \cite{Balikas2017} compares two algorithms from MMO to two QD algorithms in a robotics task, showing that clearing's performance can be comparable to that of QD. Finally, \cite{Li2019b} discusses 100 solution set quality indicators in MOO and \cite{Tian2019} discusses diversity indicators for MOO.

\section{Metrics}
\label{sec:3}

From the large number of diversity metrics available we only consider metrics that do not depend on precise domain knowledge, because no knowledge about actual local optima is available in real world applications. Three commonly used distance-based metrics are selected to evaluate the experiments in this work. 
The \textit{Sum of Distances to Nearest Neighbor (SDNN)} measures the size of a solution set as well as the dispersion between members of that set.
\textit{Solow-Polasky Diversity (SPD)} measures the effective number of species by using pairwise distances between the species in the set~\cite{Solow1994}. If the solutions are similar with respect to each other, SPD tends to 1, otherwise to $N$. The sensitive parameter $\theta$, which determines how fast a population tends to $N$ with increasing distance, needs to be parameterized for every domain. It is set to 1 for genetic distances and to 100 for phenotypic distances in this work.
\textit{Pure Diversity (PD)} is used in high-dimensional many-objective optimization~\cite{Tian2017,Wang2017b}. It does not have parameters, which makes it more robust, and depends on a dissimilarity measure ($L_{0.1}$-norm). 

Publications in the field of QD have focus on a small number of metrics. The total fitness is used directly or through the \textit{QD-score}~\cite{Pugh2015}, which calculates the total fitness of all filled niches in a phenotypic archive. To achieve this, the solutions from a non-QD algorithm are projected into a fixed phenotypic niching space. This score is domain-dependent and does not allow comparing QD algorithms that have different archiving methods. A comparison between archives created from different features introduces a bias towards one of the archives. The \textit{collection size} indicates the proportion of the niching space that is covered by the collection, but again can only be used on a reference archive~\cite{Cully2017}. Archive-dependent metrics do not generalize well and introduce biases. We therefore only use distance-based diversity metrics. The high dimensionality of phenotypic spaces is taken into account by using appropriate distance norms.

\section{Polygon Domain}
\label{sec:4}

\begin{figure}[h]
	\centering
	\includegraphics{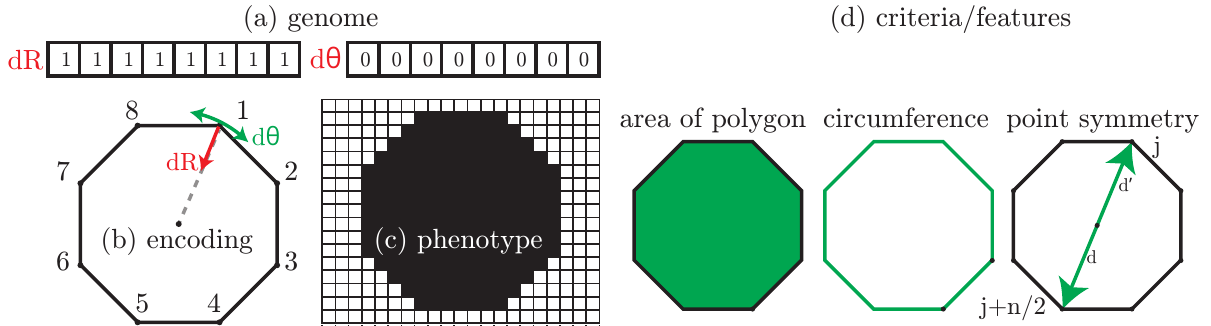}
	\caption{Free form encoding of polygons. The genome (a) consists of 16 parameters that define axial and radial deformations (b). The phenotype is considered to be the pixel representation of the polygon (c). Shown is a 20x20 phenotype, although we use 64x64 pixels. Features/criteria are shown in (d).}
	\label{img:encoding}
\end{figure}

We construct a domain of free form deformed, eight-sided polygons. The genome (Fig.~\ref{img:encoding}a) consists of 16 parameters controlling the polar coordinate deviation of the polygon control points. The first eight genes determine the deviation of the radius of the polygon's control points, the second eight genes determine their angular deviation. Since the phenotypes can be expressed as binary bitmap images (Figs.~\ref{img:encoding}b and~\ref{img:encoding}c, resolution of 64x64 pixels) we use the Hamming distance in the diversity metrics to circumvent the problem of high dimensionality~\cite{hamming1950error}.

Three aspects describing the polygons are defined that can be used either as criteria or as features (Fig.~\ref{img:encoding}d): the area of the polygon $A$, its circumference $l$ and point symmetry $P$ through the center. The polygon is sampled at $n=1000$ equidistant locations on the polygon circumference. The symmetry error $E_s$ is calculated as the sum of distances of all $n/2$ opposing sampling locations. The symmetry metric is calculated as shown in Eq.~\ref{eq}.

\begin{equation}
f_P(x_i) = {1 \over {1 + E_s(x_i)} }, E_s(x_i) = \sum_{j=1}^{n/2}||x_i^j,x_i^{j+n/2}||
\label{eq}
\end{equation}

\section{Evaluation}
\label{sec:5}

We ask which paradigm (objective space, search space or phenotype space) provides the highest phenotypic diversity of shapes. We compare VE, RLS and NSGA-II in multiple experiments. Throughout these experiments we fix the number of function evaluations and solutions and use five replicates per configuration. In NSGA-II the features are used as optimization criteria, maximizing $A$ and minimizing $l$. The true Pareto set consists of circles with varying sizes. The number of generations is set to 1024 and mutation strength to 10\% of the parameter range. The probability of crossover for NSGA-II is 90\% and probability of mutation ${1 \over dof} = 0.0625$\%, with $dof = 16$ degrees of freedom. VE's archive size is varied throughout the experiments. The number of children and population size is set to the same value. RLS uses as many restarts as the size of the VE archive, the step size is set to $\rho = 0.065$ (after a small parameter sweep) and L-BFGS-B is used as a local search method (within the bounds of the domain). The initial solution set for VE and NSGA-II is created with a Sobol sequence - the initial RLS solution is in the center of the parameter range but RLS' space filling character assures a good search space coverage. 

\subsection{Genetic or Phenotypic Diversity}

Biology has inspired evolutionary optimization to compose a solution of a genome, its encoding, and a phenotype, its expression. The phenotype often is a very high-dimensional object, for example a high-resolution 2D image, and can involve the interaction with the environment. Since the phenotypic space is usually too large, a low-dimensional representation, the genome, is used as search space. An expression function is constructed that turns a genome into its phenotype. Although the expression function should ideally be a bijective mapping, it often does not prevent multiple genomes to be mapped to the same phenotype. The phenomenon of such a surjective mapping is called genetic neutrality, which is not the same but akin to genetic neutrality in biology. In biology, a neutral mutation is understood to be a mutation that has no effect on the survivability of a life form. In evolutionary computation, genetic neutrality is referred to as genetic variants that have the same phenotype~\cite{Hu2011}.

\begin{figure}[h]
	\centering
	\includegraphics{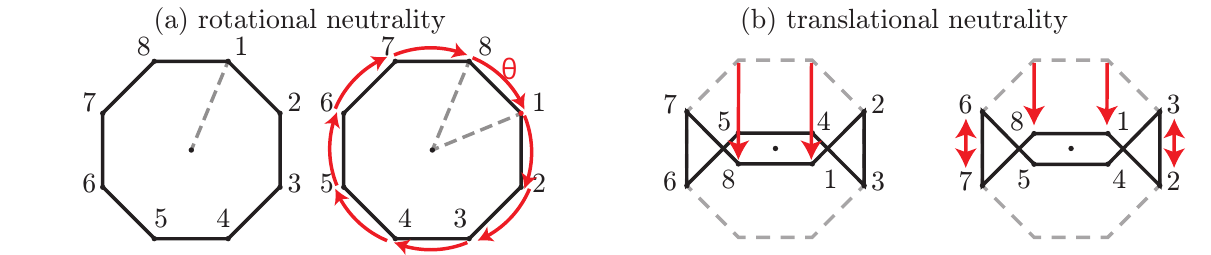}
	\caption{Genetic neutrality. The same phenotype is expressed when rotating the control points by a $\pi \over 8$ angle (left) or by translating the control points as shown (right).}
	\label{fig:neutrality}
\end{figure}

\noindent Figure~\ref{fig:neutrality}(a) shows an example polygon. If the angle $\theta$ equals 0\textdegree or 45\textdegree, phenotypically speaking, these shapes are the same. In this case, eight genomes all point to the same phenotype. Similarly, Figure~\ref{fig:neutrality}(b) shows how, through translations of the keypoints, a similar shape can appear based on different genomes. We postulate the first hypothesis: diversity maintenance in a neutral, surjective genetic space leads to lower phenotypic diversity than when using phenotypic niching.

While diversity is often thought about in terms of the distribution of points in the search space, we make a case to measure diversity in phenotypic space, which is independent of the encoding and does not suffer from the effects of genetic neutrality. Phenotypes may also include other factors that are not embodied within the solution's shape itself, but emerge through interaction with the environment. This is taken advantage of in several publications on neuroevolution~\cite{Lehman2011a,Lehman2011}. In this work we only analyse the narrow interpretation of phenotypes, which does not include behavior.

\begin{figure}[h]
	\centering
	\includegraphics{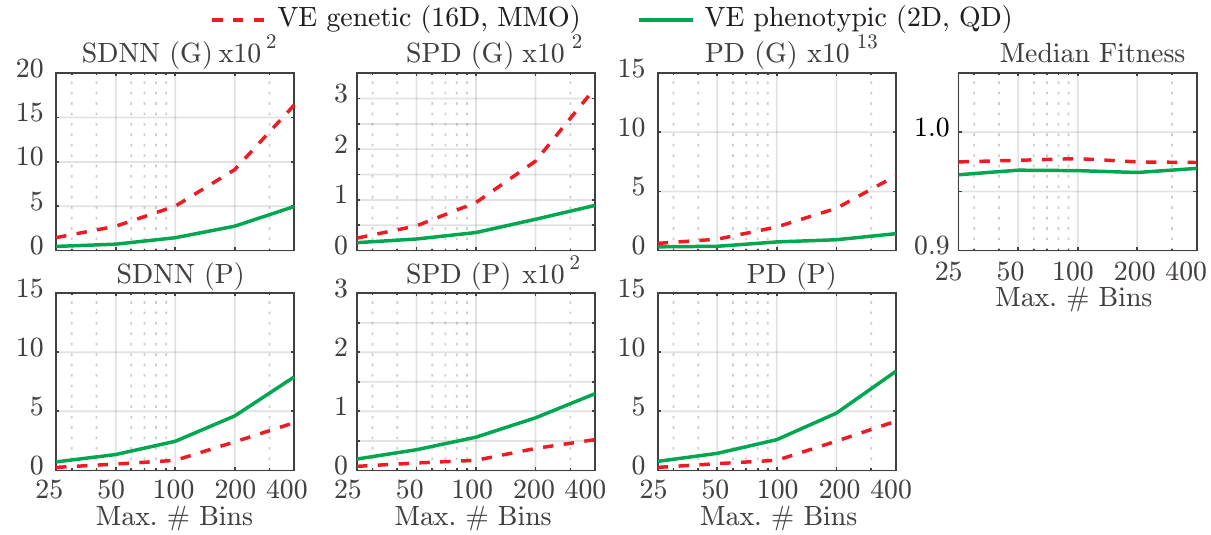}
	\caption{Voronoi-Elites (VE) performed in 16D genetic and 2D phenotypic space. Top: genetic diversity (SDNN = Sum of Distances to Nearest Neighbor, SPD = Solow-Polasky Diversity, and PD = Pure Diversity) and median fitness, bottom: phenotypic diversity. The number of bins/solutions is increased (x-axis).}		
	\label{fig:genopheno}
\end{figure}

The Voronoi tessellation used in VE makes it easy to compare archives of different dimensionality by fixing the number of niches. We apply VE as an MMO algorithm, performing niching in 16-dimensional genetic space, and as a QD algorithm with a two-dimensional phenotypic space. The number of bins is increased to evaluate when differences between genetic and phenotypic VE appear (Fig.~\ref{fig:genopheno}). At 25 solutions, the approaches produce about the same diversity, but genetic VE finds higher quality solutions. As the number of bins is increased, based on where niching is performed (genetic or phenotypic space), the diversity in that space becomes higher. Phenotypic VE beats genetic VE in terms of phenotypic diversity, which gives us evidence that the first hypothesis is valid. At the same time, the average fitness values of genetic VE are higher than that of phenotypic VE, although the difference gets lower towards 400 solutions. 

\begin{table}[h]
	\centering
	\caption{Parameter settings in order of increasing genetic neutrality.}
	\label{tbl:parameter}
	\begin{tabular}{l|l|l|l|l|l}
		case & axial min. & axial max. & radial min. & radial max. & neutrality\\
		\hline
		A & 0 		& 1 & -0.05 & 0.05 & -\\
		B & 0 		& 1 & -0.125 & 0.125& +\\
		C & -0.25 	& 1 & -0.25 & 0.25& ++\\
		D & -0.5 	& 1 & -0.5 & 0.5& +++\\
		E & -1 		& 1 & -1 & 1& ++++\\
	\end{tabular}
\end{table}

\noindent We compare phenotypic VE to NSGA-II and RLS. When we bound $dr$ between $0$ and $1$ and $d\theta$ between $+/- 0.125 \times \pi$, we can minimize genetic neutrality. Neutrality is increased by expanding those bounds (Table~\ref{tbl:parameter}). In contrast to VE, the phenotypic diversity of RLS' solutions is expected to decrease as genetic neutrality increases. Since there is no mechanism to distinguish between similar shapes with different genomes, there is an increasing probability that RLS finds similar solutions. We expect that the solution set produced by RLS due to its space filling character is more diverse than using NSGA-II. 

\begin{figure}[h]
	\centering
	\includegraphics{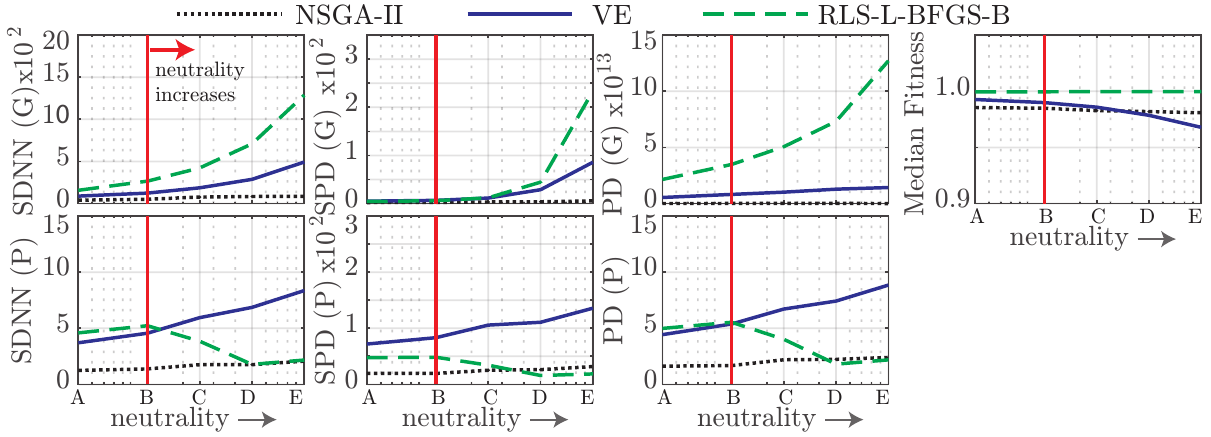}
	\caption{Genetic (top) and phenotypic (bottom) diversity, and median fitness. Right of red marker: neutrality increases, using parameter bounds shown in Table~\ref{tbl:parameter}.}	
	\label{img:stdalgos}
\end{figure}

Finally, it can make more sense to treat objectives as features and, instead of searching for the Pareto set, allowing all combinations of features and increasing the diversity of the solution set. We expect NSGA-II to easily find the Pareto set, which consists of circles of various scales, maximizing the area while minimizing the length of the circumference, while QD should find a variety of shapes that can be any combination of large and small $A$ and $l$. We postulate the second hypothesis: allowing all criteria combinations, instead of using a Pareto approach, leads to higher diversity, while still approximating the Pareto set.

The number of solutions is set to 400. A result similar to Fig.~\ref{fig:genopheno} appears for the standard algorithms in Fig.~\ref{img:stdalgos}. Phenotypic diversity is highest for VE, especially after the genetic neutrality threshold is crossed (at B). Diversity of NSGA-II is lowest, as is expected for this setup. Although diversity of VE is higher than that of RLS, the latter's solutions are all maximally symmetric (see fitness plots), making RLS much more appropriate when quality is more important than diversity. These results confirm the first part of the second hypothesis. 

\noindent The Pareto set can be calculated a priori, as we know that circular shapes maximize area while minimizing circumference. The members of the Pareto set adhere to the following genome:
$(r_1,\dots,r_8,\theta_1, \dots,\theta_8)$, where $r_i$ and $\theta_i$ have the same respective value. To create 100 shapes from the Pareto set we take ten equidistant values for $r$ and $\theta$ and combine them. 

\begin{figure}[h]
	\centering
	\includegraphics{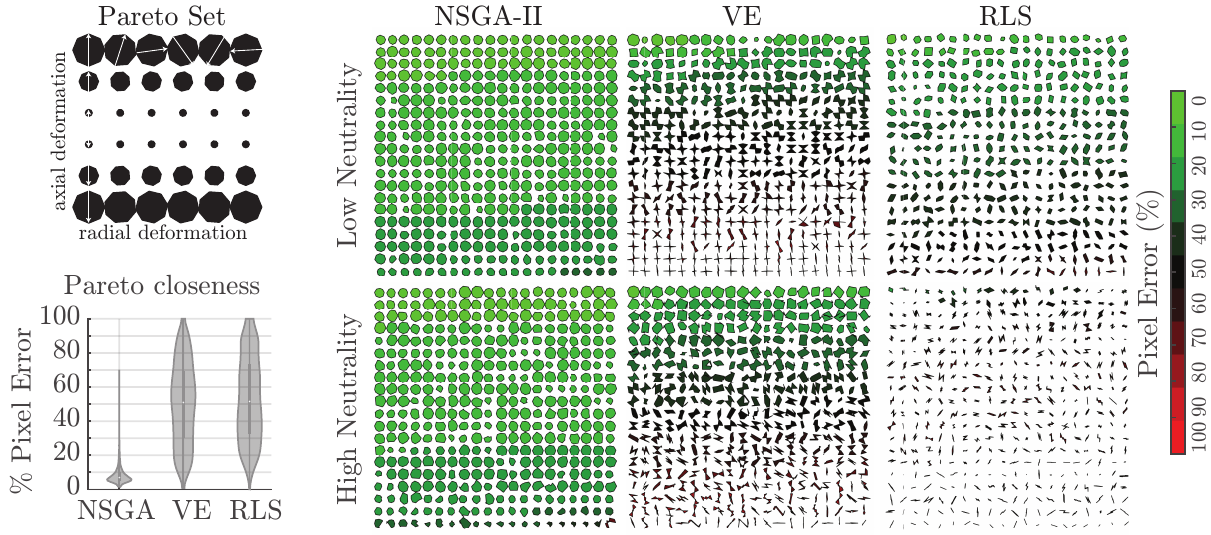}
	\caption{The ground truth Pareto set is shown over the entire parameter range, with negative as well as positive values for the radial deformation. Bottom left: closeness to Pareto set, measured as pixel errors. The six figures on the right show example solution sets for low and high neutrality.}
	\label{img:pareto}
\end{figure}

\noindent Part of the resulting Pareto set is shown in Fig.~\ref{img:pareto}. The distance to the Pareto set is measured in phenotypic space, by measuring the smallest pixel error, the sum of pixel-wise differences, between a solution and the Pareto set. We see that the a number of solutions in VE and RLS are close to the Pareto set (Fig.~\ref{img:pareto} bottom left). Example results with low and high neutrality are shown on the right. Solutions that are close to the Pareto set are shown in the brightest green color. This is evidence for the second half of the second hypothesis. VE again seems to be more robust w.r.t. genetic neutrality, as it finds more solutions close to the Pareto set in high-neutrality domains (bottom row) than RLS.

\subsection{Phenotypic Diversity without Domain Knowledge}

Up to this point we have used domain knowledge to construct a phenotypic niching space with VE. Intuitively, the area and circumference seem like good indicators for phenotypic differences. But this comparison between QD and MMO is not completely fair, as the latter does not get any domain information. On the other hand, the features used in QD might not be the most diversifying. 

\begin{figure}[h]
	\centering
	\includegraphics{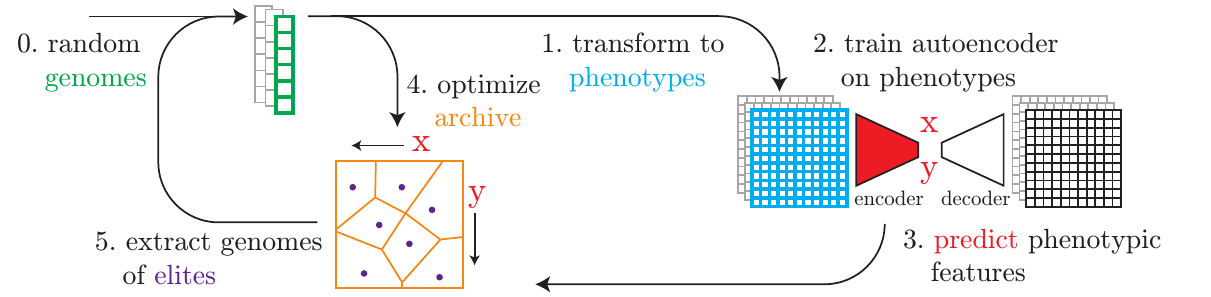}
	\caption{\textbf{AutoVE}. Generating phenotypic features with an autoencoder. A random set of genomes is created (0), their phenotypes are calculated (1) and used as a training set for an autoencoder (2). The autoencoder can now be used to predict phenotypic features of new solutions (3), which is used to fill the archive (4), after which the elite solutions are extracted from the archive (5) and used to retrain the autoencoder.}
	\label{img:autove}
\end{figure}

\noindent We remove the domain knowledge from QD and construct a phenotypic niching space by using a well known dimensionality reduction technique to map the phenotypes to a latent space, as was done in~\cite{Meyerson2016,Cully2019}. To our best knowledge, this data driven phenotypic niching approach, which we name Auto-Voronoi-Elites (AutoVE), has never been applied to shape optimization. An initial set of genomes, drawn from a quasi-random, space-filling Sobol sequence~\cite{sobol1967distribution} and expressed into their phenotypes, is used to train a convolutional autoencoder (cAE) (see Fig.~\ref{img:autove}). The bottleneck in the cAE is a compressed, latent space that assigns every phenotype to a coordinate tupel. The encoder predicts these coordinates of new shapes in the latent space, which are used as phenotypic features. QD searches phenotypes that expand and improve the cAE archive. The cAE is retrained with the new samples. The cAE consists of two convolutional layers in the encoder and four transposed convolutional layers in the decoder. We set the filter size to three pixels, the stride to two pixels, and the number of filters to eight. The cAE is trained using ADAM~\cite{kingma2015adam} with a learning rate of 0.001 and 350 training epochs and a mean square error loss function. Latent coordinates are normalized between 0 and 1. The number of generations (1024) is divided over two iterations for AutoVE and the number of latent dimensions is set to two (to compare with manual VE), five or ten. 

\begin{figure}[h]
	\centering
	\includegraphics{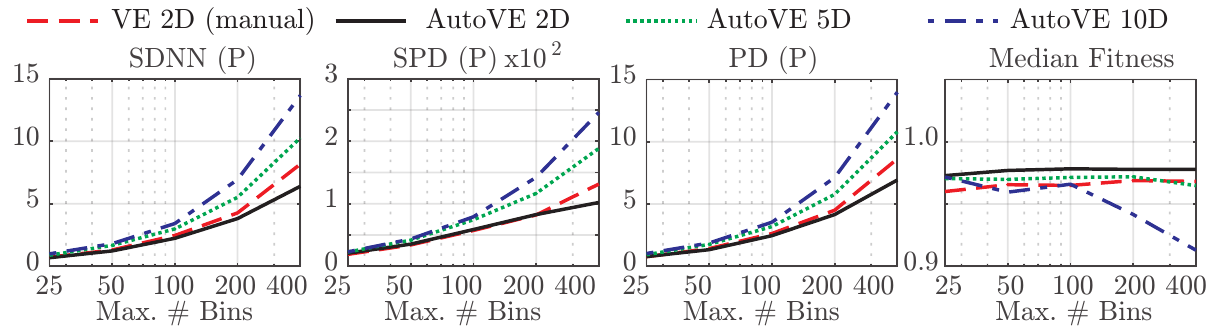}
	\caption{Phenotypic diversity and fitness of manually crafted features (VE) compared to using an autoencoder (AutoVE) with 2, 5 or 10 latent dimensions.}
	\label{fig:auto}
\end{figure}

\noindent Fig.~\ref{fig:auto} shows that the two-dimensional manual and autoencoded phenotypic space (AutoVE 2D) produce similar diversity, whereby the quality of solutions from AutoVE 2D is higher. The higher-dimensional latent spaces increase the solution set diversity at the cost of fitness. This is to be expected, as lower-fitness optima are protected in their own niches. Finally, the diversity of higher-dimensional AutoVE is around 50\% higher than any of the other tested methods.

\section{Conclusion}
\label{sec:6}

The main contributions of this work are as follows: a domain was introduced that allows comparing three different diversity paradigms; a case was made to measure diversity in phenotypic rather than genetic space; the hypothesis that QD is less sensitive to genetic neutrality than MMO was confirmed; the hypothesis that while the diversity of solutions sets of QD and RLS is higher than that of MOO, they also find some solutions close to the ground truth Pareto set, was confirmed; we showed that phenotypic diversity in QD is higher than MMO and MOO. Furthermore, we introduced VE, a simpler and self-expanding version of QD. We also used an autoencoder to discover phenotypic features in a shape optimization problem, showing that we do not need to manually predefine features to get a highly diverse solution set, allowing us to fairly compare QD to MOO and MMO. Using an autoencoder produces higher diversity than manually defined features, making AutoVE a strong choice for high diversity multi-solution optimization.

Since all paradigms have their strengths and weaknesses, we propose a guide for when to use which approach. MOO should be used when you want to optimize all the criteria and want to know the trade-off solutions between those criteria. MMO is appropriate when you have a non-neutral bijective encoding, when you have a single criterion you want to optimize for or if you want to perform a gradient-based, Quasi-Newton or (direct) evolutionary local search to refine local optima. We cannot easily do this in QD due to the effect of neutrality that allows a search to ``jump out of'' a phenotypic niche. QD should be used if you have some criteria where you are less determined about whether to optimize for them, for example during the first phase of a design process. Some representatives from the Pareto set will still be discovered. When you are interested in the largest diversity of solutions and are more willing to get some solutions with lower fitness than when using MMO, QD is the better alternative. One of the biggest strengths of QD is the possibility to understand relationships between features or even to discover features automatically.

Some research effort should be focused on hybridization. MOO and QD are connected, as the boundary of valid solutions in the phenotypic archive is close to the Pareto front, yet there is room for improvement. Connecting MMO and QD means to use a local search method in QD, which needs to overcome the genetic neutrality problem. We cannot search close to a solution in genetic space and expect newly created solutions to be close in phenotypic space. 

We gave insights about different variations of diversity and when and where to apply them, depending on whether one is most interested in trade-offs between criteria, increasing diversity while maximizing fitness, or maximizing diversity while finding high-performing solutions in a manually defined or automatically extracted phenotypic space. It is often easy to manually define two or three phenotypic descriptors, but human imagination can run out of options quickly. Automatic discovery of phenotypic features is a more attractive option for increasing solution diversity. Real world multi-solution optimization and understanding solution diversity are important steps towards increasing the efficacy and efficiency at which engineers solve problems.

%

\bibliographystyle{splncs04}
\bibliography{references}
\end{document}